\documentclass[11pt,a4paper]{article}
\usepackage[hyperref]{acl2019}
\usepackage{times}
\usepackage{latexsym}
\usepackage{multirow}
\usepackage{makecell}

\usepackage{url}

\aclfinalcopy % Uncomment this line for the final submission
\title{FinBERT: A Pretrained Language Model for Financial   Communications}

\author{\makecell{Yi Yang~~~~~~~~Mark Christopher Siy UY ~~~~~~~~Allen Huang} \\
School of Business and Management, Hong Kong University of Science and Technology\\
{\tt $\{$imyiyang,acahuang$\}$@ust.hk}, ~~~ {\tt mcsuy@connect.ust.hk} \\}
  
\date{}

\begin{document}
\maketitle
\begin{abstract}

Contextual pretrained language models, such as BERT \citep{devlin2019bert},  have made significant breakthrough in various NLP tasks by training on large scale of unlabeled text resources. Financial sector also accumulates large amount of financial communication text. However, there is no pretrained finance specific language models available.  In this work, we address the need by pretraining a financial domain specific BERT models, FinBERT, using a large scale of financial communication corpora. Experiments on three financial sentiment classification tasks confirm  the advantage of FinBERT over generic domain BERT model.
The code and pretrained models are available at \url{https://github.com/yya518/FinBERT}. We hope this will be useful for practitioners and researchers working on financial NLP tasks.

%This document contains the details of training a Business-BERT model. we address this need by exploring and releasing BERT models for finanical text.
\end{abstract}

\section{Introduction}

The growing maturity of NLP techniques and resources is drastically changing the landscape of finanical domain. %s, and financial domain is one of such emerging ground. 
%as a primary ground for the NLP techniques, and 
Capital market practitioners and researchers have keen interests in using  NLP techniques to monitor market sentiment in real time from online news articles or social media posts, since sentiment can be used as a directional signal for trading purposes. Intuitively, if there is positive information about a particular company, we expect that company's stock price to increase, and vice versa. For example, Bloomberg, the financial media company, %has applied sentiment analysis techniques to assign a sentiment score to a news story or tweet. Bloomberg 
reports that trading sentiment portfolios outperform the benchmark index significantly \citep{cui2016embedded}.  Prior financial economics research also reports that news article and social media sentiment could be used to predict market return and firm performance \citep{tetlock2007giving,tetlock2008more}. 

%Rapid development of natural language processing technologies and access to cheap computational capacity have paved the way for automatic sentiment analysis and the emergence of robo-readers in computational finance. However, the technology is still in its nascent state. 
%
Recently, unsupervised pre-training of language models on large corpora has significantly improved the performance of many NLP tasks. The language models are pretained on generic corpora such as Wikipedia. However, sentiment analysis is a strongly domain dependent task.   Financial sector  has accumulated large scale of text of financial and business communications. Therefore, leveraging the success of unsupervised pretraining and large amount of financial text could potentially benefit wide range of financial applications. 

To fill the gap,  we pretrain \textbf{FinBERT}, a finance domain specific BERT model on a large financial communication corpora of 4.9 billion tokens, including corporate reports, earnings conference call transcripts and analyst reports. We document the financial corpora and the FinBERT pretraining details. Experiments on three financial sentiment classification tasks shows that FinBERT outperforms the generic BERT models. % that trains on generic text. 
Our contribution is straightforward: we compile a large scale of text corpora that are the most representative in financial and business communications. We pre-train and release FinBERT, a new resource demonstrated to improve performance on financial sentiment analysis.

%annotated collections are a scarce resource and many are reserved for proprietary use only
% Furthermore, the lack of readily available corpus in the financial domain makes this endeavor considerably harder. Nonetheless, the creation of such a Finance-BERT could have great use in future academic work, and in the industry by different companies. 

%Contextual pretrained language model has made breakthrough in NLP, such as BERT. However, there is no pre-trained finance-specific BERT model available. We address this need by training a finance-specific BERT model using large scale of corporate-finance related text data. We document the training details and corpus.  We also compare the trained finance-specific BERT model with original BERT model on several tasks, including return prediction, volatility prediction, sentiment prediction. We release the code and pretrained models to the large audience who is doing NLP on financial tasks.

\section{Related Work}
Recently, unsupervised pre-training of language models on large corpora, such as BERT \citep{devlin2019bert}, ELMo \citep{peters2018elmo}, ULM-Fit \cite{howard2018universal}, XLNet, and GPT \citep{radford2019language} has significantly improved performance on many natural language processing tasks, from sentence classification to question answering. Unlike traditional word embedding \citep{word2vec,pennington2014glove} where word is represented as a single vector representation, these language model returns contextualized embeddings for each word token which can be fed into downstream tasks. 

The released language models are trained on general domain corpora such as news articles and Wikipedia. Even though it is easy to fine tune the language model using downstream task, it has been shown that pre-training a language model using large-scale domain corpora can further improve the task performance than fine-tuning the generic language model. To this end, several domain-specific BERT models are trained and released. 
BioBERT \cite{biobert} pretrains a biomedical domain-specific language representation model using large-scale biomedical corpora. %It shows that BioBERT outperforms BERT and previous state-of-the art models in a variety of biomedical text mining tasks. 
Similarly, ClinicalBERT \cite{clinicalbert-notes} applies BERT model to clinical
notes for hospital readmission prediction task, and \cite{clinicalBert-summaries} applies BERT on clinical notes and discharge summaries.  SciBERT \cite{SciBERT} trains a scientific domain-specific BERT model using a large multi-domain corpus of scientific publications to improve performance on downstream scientific NLP tasks. 
%Financial domain also accumulated large scale of financial communication in the written text. 
We are the first to pre-train and release a finance domain specific BERT model. 

\section{Financial Corpora}
We compile a large financial domain corpora that are most representative in finance and business communications.%: Corporate Annual/Quarterly Report 10-K \& 10-Q, Earnings  Call Transcripts, and Analyst Reports. 

\noindent\textbf{{Corporate Reports 10-K \& 10-Q}} The most important text data in finance and business communication is corporate report.
In the United States, the Securities Exchange Commission (SEC) mandates all publicly traded companies to file annual reports, known as Form 10-K, and quarterly reports, known as Form 10-Q. This document provides a comprehensive overview of the company's business and financial condition. Laws and regulations prohibit companies from
making materially false or misleading statements in
the 10-Ks. The Form 10-Ks and 10-Qs are publicly available and can be accesses from SEC website.\footnote{http://www.sec.gov/edgar.shtml}

We obtain 60,490 Form 10-Ks and 142,622 Form 10-Qs of Russell 3000 firms during 1994 and 2019 from SEC website. %We choose Russell-3000  constituent firms for reasons of importance and tractability. Firms in the Russell-3000  index represents approximately 98\% of the American public equity market capitalization. 
We only include sections that are textual components, such as Item 1 (Business) in 10-Ks, Item 1A (Risk Factors) in both 10-Ks and 10-Qs and Item 7 (Management’s Discussion and Analysis) in 10-Ks.
%For Form 10-Ks, we include three sections: Item 1 (Business), Item 1A (Risk Factors) and Item 7 (Management’s Discussion and Analysis). For Form 10-Qs, we include two sections: Item 2 (Management’s Discussion and Analysis) and Item 1A (Risk Factors). We choose to include the above sections since they are the  textual components in corporate reports.

\noindent\textbf{Earnings Call Transcripts}
Earnings calls are quarterly conference calls that company executives hold with  investors and analysts to discuss firm overall performance. During an earnings call, executives such as CEOs and CFOs read forward-looking statements and provide their information and interpretation of their firm’s performance during the quarter. Analysts also have the opportunity to request managers to clarify information. % and solicit additional information that the management team does not disclose in the statement. %The National Investor Relations Institute reports that 92\% of companies conduct earnings calls. 
Institutional and individual investors listen to the earnings call and spot the tones of executives that portend good or bad news for the company. We obtain 136,578 earnings conference call transcripts of 7,740 public firms between 2004 and 2019.  The earnings call transcripts are obtained from the website \texttt{Seeking Alpha}\footnote{https://seekingalpha.com/}.

\noindent\textbf{Analyst Reports} Analyst reports are another useful source of information for institutional and individual investors \citep{sri1987investor}. An analyst report typically provides several quantitative summary measures, including a stock recommendation, an earnings forecast, and sometimes a target price. It also provides a detailed, mostly textual analysis of the company. Institutional investors spend millions of dollars annually to purchase the full content of analyst reports to read the written textual analysis. We obtain analyst reports in the Investext database issued for S\&P firms during the 1995-2008 period, which yields  a set of 488,494 reports.

\noindent\textbf{Overall Corpora Statistics}
%All data are processed into the required format following BERT's requirement. 
The total size of all 4 corpora is approximately  4.9 billion tokens. We present the pretraining financial corpora statistics in Table\ref{tab:stats}. As a comparison, BERT's pre-training corpora consists of two textual corpora with a total of 3.3 billion tokens. % BERT model requires all input sentences to follow a specific format. Particularly, each line in the input file represents a single sentence, and documents are separated via empty lines. %In order to do this, sentence segmentation was carried out on the data using Spacy's existing off the shelf tool. Furthermore, reading the data from the raw input file proved to be troublesome without using the right codec. For this data set: \textit{"unicode-escape"} was utilized. The presence of headers in the data set proved to be an interesting discussion point as they are not technically sentences. However, it was decided that they should be included as they still provide some valuable semantic and structural information about the document. 
%After all data are processed into the required format, they are then used to create new pretraining data, using the $\textit{create\_pretraining}$ script available at the original Bert's github repository. The hyper parameters uses were all the default values on the original Bert Repo. 

% Please add the following required packages to your document preamble:
\begin{table}[h]
\centering
\begin{tabular}{ll}
\hline
 Corpus & \# of tokens \\ \hline
%\multirow{2}{*}{BERT} & English Wiki & 2.5B \\
% & BookCorpus & 0.8B \\ \hline
 Corporate Reports 10-K \& 10-Q &  2.5B \\
  Earnings Call Transcripts & 1.3B \\
  Analyst Reports & 1.1B \\ \hline
\end{tabular}
\caption{Size of pretraining financial corpora.}
\label{tab:stats}
\end{table}
\section{FinBERT Training}

\noindent\textbf{Vocabulary}
We construct FinVocab, a new Word-Piece vocabulary on our financial corpora using the SentencePiece library. We produce both cased and uncased versions of FinVocab, with sizes of 28,573 and 30,873 tokens respectively. This is very similar to the 28,996 and 30,522 token sizes of the original BERT cased and uncased BaseVocab. The resulting overlap between between the original BERT BaseVocab, and FinVocab is  41\% for both the cased and uncased versions. %This illustrates a large difference in the frequently used words between financial and general domain texts. 

\noindent\textbf{FinBERT-Variants}
We use the original BERT code \footnote{\url{https://github.com/google-research/bert}} to train FinBERT on our financial corpora with the same configuration as BERT-Base. Following the original BERT training, we set a maximum sentence length of 128 tokens, and train the model until the training loss starts to converge. We then continue training the model allowing sentence lengths up to 512 tokens. 
In particular, we train four different versions of FinBERT: cased or uncased; BaseVocab or FinVocab. 

%In this section, we provide training details for the finance-BERT model. The training method follows the method specified in the original BERT Tensorflow Repo. Instead of training the BERT model from scratch, we instead do additional pre-training on top of the released BERT model, which was originally pre-trained on a general corpus for approximately one million epochs. However, now the pre-training is done using the corpus discussed in part 3.  This follows the work done by other papers, who try to train their own domain-specific BERT, such as [], and [].

%We train four BERT models on financial corpora. 

%\begin{itemize}

\textbf{{FinBERT-BaseVocab, uncased/cased:}} Model is initialized from the original BERT-Base uncased/cased model, and is further pretrained on the financial corpora for 250K iterations at a smaller learning rate of $2e^{-5}$, which is recommended by BERT code. 

%\item \textbf{{FinBERT, BaseVocab, cased}}. Model is initialized from the original BERT-Base cased model, and is further pretrained on the financial corpora for 250K iterations at a smaller learning rate. 

\textbf{{FinBERT-FinVocab, uncased/cased:}} Model is trained from  scratch using a new uncased/cased financial vocabulary FinVocab for 1M iterations.

%\item \textbf{{FinBERT, FinVocab, uncased }}. Model is trained from scratch using a new uncased financial vocab, derived from the financial corpora, for 1M iterations. 

%\end{itemize}

\noindent\textbf{Training} The entire training is done using a NVIDIA DGX-1 machine. The server has 4 Tesla P100 GPUs, providing a total of 128 GB of GPU memory. This machine enables us to train the BERT models using a batch size of 128. We utilize Horovord framework \citep{sergeev2018horovod} for multi-GPU training. %The total number of additional epochs steps for which the BERT model is pre-trained for is 250,000. 
%We monitor the loss function and do not observe overfitting issues. 
Overall, the total time taken to perform pretraining for one model is approximately 2 days.
With the release of FinBERT, we hope  financial  practitioners  and  researchers  can benefit from FinBERT model without the necessity of the significant computational resources required  to  train  the  model.   
%We hope releasing our pretrained models will be useful to the community.% This does not count the time taken for CPU expensive data prepossessing step. 

% \begin{table*}[t]
% \centering
% \begin{tabular}{ll}
%   & All \\ \hline
% Tf-idf with Naive Bayes &  67\%\\
% Word2vec  &  51\%\\
% BERT &  84\%\\
% FinanceBERT  & 87\%\\
% \hline
% \end{tabular}
% \caption{Sentiment Classification Performance for different baselines and FinanceBERT.}
% \label{tab:sentiment}
% \end{table*}

% \begin{table*}[t]
% \centering
% \begin{tabular}{lllll}
%  & 10-K\&10-Q & Conference Call & Analyst Report & All \\ \hline
% Tf-idf with Naive Bayes &  &  &  &  \\
% Word2vec &  &  &  &  \\
% BERT &  &  &  &  \\
% FinanceBERT &  &  &  & \\
% \hline
% \end{tabular}
% \caption{Sentiment Classification Performance for different baselines and FinanceBERT.}
% \label{tab:sentiment}
% \end{table*}
\section{Financial Sentiment Experiments}
Given the importance of sentiment analysis in financial NLP tasks, we conduct experiments on financial sentiment classification datasets.

\subsection{Dataset}

\noindent\textbf{{Financial Phrase Bank}} is a public dataset for financial sentiment classification \citep{malo2014good}. The dataset contains 4,840 sentences selected from financial news. The dataset is manually labeled by 16 researchers with adequate background knowledge on financial markets. The sentiment label is either positive, neutral or negative.

\noindent \textbf{{AnalystTone Dataset}} is a dataset to gauge the opinions in analyst reports, which is commonly used in Accounting and Finance literature \citep{huang2014evidence}. The dataset contains randomly selected 10,000 sentences from analyst reports in the Investext database. Each sentence is manually annotated into one of three categories: positive, negative and neutral. This classification yields a total of 3,580 positive, 1,830 negative, and 4,590 neutral sentences in the  dataset.

\noindent\textbf{{FiQA Dataset}} is an open challenge dataset for  financial sentiment analysis, containing 1,111 text sentences \footnote{\url{https://sites.google.com/view/fiqa/home}}. Given an English text sentence in the financial domain (microblog message, news statement), the task of this challenge is to predict the associated numeric sentiment score, ranged from -1 to 1. We convert the original regression task into a binary classification task for consistent comparison with the above two datasets. %Since the sentiment score is numerical, the task is evaluated using root mean square error (RMSE). %Using the provided sentiment scores, this regression task can also be turned into a binary classification task.

%, negative, if the score is less than or equal to 0, and positive, if the score is greater than 0. A neutral classification for sentences with a sentiment score of 0 was not considered because there are only 12 such sentences in the dataset. 

We randomly split each dataset into 90\% training and 10\% testing 10 times and report the average. Since all dataset are used for sentiment classification, we report the accuracy metrics in the experiments. 
%We split the Analyst Opinion and Financial Phrase Bank datasets into 90\% training and 10\% testing. While, for the FiQA dataset, we use a 85\% training and 15\% testing split since it is a much smaller dataset. 

\subsection{Fine-tune Strategy}
We follow the same fine-tune architecture and optimization choices used in \citep{devlin2019bert}. We use a simple linear layer, as our classification layer, with a softmax activation function. We also use cross-entropy loss as the loss function. %For the regression task, we use the same architecture as in the classification task, minus the activation function. The loss function used to train the model also changes to L2-loss, mean square error. For either task, %The learning rate used for updating the parameters of the linear layer is 1e-3, while the learning rate used for the updating the parameter value of the BERT architecture is 1e-5. 
Note that an alternative is to feed the contextualized word embeddings of each token into a deep architectures, such as Bi-LSTM, atop frozen BERT embeddings. We choose not to use this strategy as it has shown to perform significantly worse than fine-tune BERT model \citep{SciBERT}.

\subsection{Experiment Results}
We compare FinBERT with original BERT-Base model \citep{devlin2019bert}, and we evaluate both cased and uncased versions of this model. The main results of financial sentiment analysis tasks are present in Table \ref{tab:main}.

\begin{table*}[]
\centering
\begin{tabular}{ccccccc}
\hline
 & \multicolumn{2}{c}{BERT} & \multicolumn{2}{c}{FinBERT-BaseVocab} & \multicolumn{2}{c}{FinBERT-FinVocab} \\ \hline
 & \multicolumn{1}{c}{cased} & \multicolumn{1}{c}{uncased} & \multicolumn{1}{c}{cased} & \multicolumn{1}{c}{uncased} & \multicolumn{1}{c}{cased} & \multicolumn{1}{c}{uncased} \\ \hline
PhraseBank & 0.755 & 0.835 & 0.856 & 0.870 & 0.864 & \textbf{0.872}  \\
%FiQA  & 0.148 & 0.128  & 0.121 & 0.108 & 0.118 & \textbf{0.100} \\
FiQA & 0.653  & 0.730  & 0.767 & 0.796 & 0.814 & \textbf{0.844} \\
AnalystTone & 0.840  & 0.850  & 0.872 & 0.880 & 0.876  & \textbf{0.887} \\ \hline
\end{tabular}
\caption{Performance of different BERT models on three financial sentiment analysis tasks.}
\label{tab:main}
\end{table*}

% \begin{table*}[]
% \centering
% \begin{tabular}{ccccc}
% \hline
%  & 10-Ks/10-Qs & Earnings Call  & Analyst Reports & All \\ \hline
% PhraseBank & 0.847 & 0.860 & 0.861 & 0.864  \\
% FiQA & 0.125  & 0.123  & 0.120  & 0.118 \\
% AnalystTone & 0.858 & 0.870 & 0.872 & 0.876  \\ \hline

% \end{tabular}
% \caption{Performance on different financial corpora.}
% \label{tab:my-table}
% \end{table*}

\begin{table*}[]
\small
\centering
\begin{tabular}{cccccc}
\hline
 &  & 10-Ks/10-Qs & Earnings Call & Analyst Reports & All \\ \hline
\multicolumn{1}{c|}{} & PhraseBank & 0.835 & 0.843 & 0.845 & 0.856 \\
%\multicolumn{1}{c|}{BaseVocab} & FiQA & 0.133 & 0.1296 & 0.126 & 0.121 \\
\multicolumn{1}{c|}{BaseVocab} & FiQA  & 0.707 & 0.731 & 0.744 & 0.767 \\
\multicolumn{1}{c|}{} & AnalystTone & 0.845 & 0.862 & 0.871 & 0.872  \\ \hline
\multicolumn{1}{c|}{} & PhraseBank & 0.847 & 0.860 & 0.861 & 0.864 \\
%\multicolumn{1}{c|}{FinVocab} & FiQA & 0.125 & 0.123 & 0.120 & 0.118 \\
\multicolumn{1}{c|}{FinVocab} & FiQA  & 0.766  & 0.778  & 0.796 & 0.814 \\
\multicolumn{1}{c|}{} & AnalystTone & 0.858 & 0.870 & 0.872 & 0.876 \\ \hline
\end{tabular}
\caption{Performance of pretraining on different financial corpus.}
\label{tab:table2}
\end{table*}

\noindent \textbf{FinBERT vs. BERT}
The results show substantial improvement of FinBERT models over the generic BERT models. On PhraseBank dataset, the best model uncased FinBERT-FinVocab achieves the accuracy of 0.872, a 4.4\% improvement over uncased BERT model and 15.4\% improvement over cased BERT model. On FiQA dataset, the best model uncased FinBERT-FinVocab achieves the accuracy of 0.844, a 15.6\% improvement over uncased BERT model and a 29.2\% improvement over cased BERT model. Lastly, on the AnalystTone dataset, the best model uncased FinBERT-FinVocab improves the uncased and cased BERT model by 4.3\% and 5.5\% respectively. Overall speaking, pretraining on financial corpora, as expected, is effective and enhances the downstream financial sentiment classification tasks. In financial markets where capturing the accurate sentiment signal is of utmost importance, we believe the overall FinBERT improvement demonstrates its practical utility. 

\noindent \textbf{FinVocab vs. BaseVocab}
We assess the importance of an in-domain financial vocabulary by pre-training different FinBERT models using BaseVocab and FinVocab. For both uncased and cased model, we see that FinBERT-FinVocab outperforms its BaseVocab counterpart. However, the performance improvement is quite marginal on PhraseBank and AnalystTone task. Only do we see substantial improvement on FiQA task (0.844 vs. 0.796). Given the magnitude of improvement, we suspect that while an in-domain vocabulary is helpful, FinBERT benefits most from the financial  communication corpora pretraining. 

\noindent \textbf{Cased vs. Uncased} We follow \citep{devlin2019bert} in using both the cased model and the uncased model for all tasks. Experiments result suggest that uncased models perform better than cased models in all tasks. This result is consistent with prior work of Scientific domain and Biomedical domain BERT models.

\noindent \textbf{Corpus Contribution}
%In addition to the FinBERT model trained on all three corpora, 
We also train different FinBERT models on three financial corpus separately. The performance of different FinBERT models (cased version) on different tasks are present in Table \ref{tab:table2}. It shows that FinBERT trained on all corpora  achieves the overall best performance indicating that combining additional financial communication corpus could improve the language model quality. Among three datasets, Analyst Reports dataset appears to perform well in three different tasks, even though it only has 1.1 billion word tokens.  Prior research finds that corporate report such as 10-Ks and 10-Qs contains  redundant content, and that a substantial amount of textual volume contained in 10-K reports is attributable to managerial discretion in how firms respond to mandatory disclosure requirements \citep{cazier201610}.  Does it suggest that Analyst Reports data  contains more information content than corporate reports and earnings call transcripts? We leave it for future research.
\section{Conclusion}

In this work, we pre-train a financial-task oriented BERT model, FinBERT. The FinBERT model is trained on a large  financial corpora that are representative of English financial communications. We show that FinBERT outperforms generic BERT models on three financial sentiment classification tasks.
%We choose sentiment classification as our evaluation task because various research has shown market sentiment can be used as trading signals for financial markets. 
% we hope financial practitioners and researchers can benefit from FinBERT model without the necessity of the significant computational resources required to train the model.
With the release of FinBERT, we hope practitioners and researchers can utilize FinBERT for a wider range of applications where the prediction target goes beyond sentiment, such as financial-related outcomes including stock returns, stock volatilities, corporate fraud, etc.

%We evaluate the FinBERT performance on several financial tasks, return prediction, volatility prediction and sentiment classification. FinBERT achieves the best performance on all tasks.  To the best of our knowledge, our work is the first to release financial domain-specific BERT model. 

%We hope that our released FinBERT model can bridge the gap between financial applications and natural language understanding. More importantly, our hope is that all capital market participants will be able to benefit from FinBERT model without the necessity of the significant computational resources required to train the model.

\section*{Acknowledgments}
This work was supported by Theme-based Research Scheme (No. T31-604/18-N) from Research Grants Council in Hong Kong. %We thank the anonymous reviewers for helpful comments. Any opinions, findings, conclusions, or recommendations expressed here are those of the authors and do not necessarily reflect the view of the sponsor.

\bibliographystyle{acl_natbib}
\bibliography{ref}

\appendix

\end{document}